\documentclass[11pt,twoside]{article}

\usepackage{amsmath}
\usepackage{amssymb}
\usepackage{fancyhdr}
\usepackage{graphicx}
\usepackage{subfigure}
\usepackage[textwidth=16cm,textheight=22cm,top=3cm,bottom=3cm]{geometry}

\newtheorem{theorem}{Theorem}
\newtheorem{remark}{Remark}

\newcommand{\drop}[1]{}
\newcommand{\no}{\noindent}
\newcommand{\fer}[1]{(\ref{#1})}
\newcommand{\qtext}[1]{\quad\text{#1}}

\newcommand{\bx}{\mathbf{x}}
\newcommand{\by}{\mathbf{y}}

\newcommand{\cK}{\mathcal{K}}

\newcommand{\N}{\mathbb{N}}
\newcommand{\R}{\mathbb{R}}

\def\O{\Omega}

\newcommand{\abs}[1]{| #1 |}
\newcommand{\nor}[1]{\| #1 \|}

\DeclareMathOperator{\cum}{cum}

\DeclareMathOperator{\NF}{NF}

\title{On a new formulation of nonlocal image filters involving the relative rearrangement
\thanks{Supported by Spanish MCI Project MTM2010-18427.}}

\author{Gonzalo Galiano  \thanks{Dpt. of Mathematics, Universidad de Oviedo,
 c/ Calvo Sotelo, 33007-Oviedo, Spain ({\tt galiano@uniovi.es, julian@uniovi.es})}
    \and Juli\'an Velasco\footnotemark[2] }
\date{}

\begin{document}

\maketitle
% \footnote[]{\emph{Keywords: }neighborhood filters, decreasing rearrangement, denoising, segmentation}
% \footnote{\emph{2010 Mathematics Subject Classification: }68U10}
\begin{abstract}
Nonlocal filters are simple and powerful techniques for image denoising.
In this paper we study the reformulation of a broad class of nonlocal filters
in terms of two functional rearrangements: the decreasing and the relative 
rearrangements.

Independently of the dimension of the \emph{image}, we reformulate 
these filters as integral operators defined in a one-dimensional space
corresponding to the level sets measures.

We prove the equivalency between the original and the rearranged versions of the
filters and propose a discretization in terms of constant-wise interpolators, 
which we prove to be convergent to the solution of the continuous setting.

For some particular cases, 
this new formulation allows us to perform a detailed analysis of the filtering properties.
Among others, we prove that the filtered image is a contrast change of the original image, and that the filtering procedure behaves asymptotically as a 
shock filter combined with a border diffusive term, responsible for 
the staircaising effect and the loss of contrast.

\vspace{0.25cm}

\no\emph{Keywords: }Nonlocal image filters, Neighborhood filter, Bilateral filter, decreasing rearrangement, relative rearrangement, denoising.

%\subclass{68U10}
\end{abstract}

%\pagestyle{myheadings}
%\thispagestyle{plain}
%\markboth{GONZALO GALIANO AND JULIAN VELASCO}{NEIGHBORHOOD FILTERS AND THE DECREASING REARRANGEMENT}

%%%%%%%%%%%%%%%%%%%%%%%%%%%%%%%%%%%%%%%%%%%%%%%%%%%%%%
\section{Introduction}

Let $\O\subset\R^d$ $(d\geq 1)$ be an open and bounded set,
$u\in L^\infty(\O)$, and consider the nonlocal family of filters defined by
\begin{align}
 \label{def.NL}
   \text{F}_{h} u(\bx)= 
   \frac{1}{C(\bx)}\int_\O \cK_h(u(\bx)-u(\by)) w (\bx,\by)  u(\by)d\by,
 \end{align}
 where $h$ is a positive constant, and $C(\bx)=\int_\O \cK_h(u(\bx)-u(\by)) w (\bx,\by) d\by$ is a normalization factor. 
 
Functions $\cK_h(\xi)=\cK(\xi/h)$ and $w$ are the kernels of the filter. A usual choice for $\cK$ is the Gaussian $\cK(\xi)=\exp(-\xi^2)$, while different choices of $w$ give rise to several well known nonlocal filters, e.g.,  
\begin{itemize}
 \item The Neighborhood filter, see \cite{Buades2005}, for $w(\bx,\by)\equiv 1$.
 \item The Yaroslavsky filter \cite{Yaroslavsky1985,Yaroslavsky2003}, 
 for $w(\bx,\by)\equiv \chi_{B_\rho(\bx)}(\by)$, the characteristic function of a ball centered at $\bx$ of radious $\rho>0$.
 \item The SUSAN \cite{Smith1997} and Bilateral filters \cite{Tomasi1998}, 
 for \[w(\bx,\by)=\text{e}^{-\frac{\abs{\bx-\by}^2}{\rho^2}},  \quad  \rho>0.\]
 
%  \item The cross or joint Bilateral filter \cite{eisemann2004,petschnnig2004}, 
%  with, for some function $\bar w$ usually related to the filtered image $u$,  
%  \[ w(\bx,\by)= \text{e}^{-\frac{\abs{\bx-\by}^2}{h^2}} \frac{\bar w(\by)}{u(\by)},\quad
%  w(\bx,\by)= \text{e}^{-\frac{\abs{\bx-\by}^2}{h^2}} , \] and its corresponding cross or joint Neighborhood filter, for $w(\bx,\by)= \bar w(\by)/u(\by)$, $w(\bx,\by)\equiv 1$.
 
 \item The weighted Bilateral filter \cite{matsuo2013}, 
 with, for some function $\bar w$ usually related to the depth map of $u$,  
 \[ w(\bx,\by)= \text{e}^{-\frac{\abs{\bx-\by}^2}{\rho^2}} \bar w(\by), \quad  \rho>0,\] and its corresponding weighted Neighborhood filter, for $w(\bx,\by)= \bar w(\by)$.

\end{itemize}

 %Nonlocal Means filter (NLM) \cite{Buades2005} by Buades, Coll and Morel.

These filters have been introduced in the last decades
as efficient alternatives to local methods such as those expressed in terms 
of nonlinear diffusion partial differential equations (PDE's), among which the pioneering approaches of Perona and Malik \cite{Perona1990}, \'Alvarez, Lions and Morel \cite{Alvarez1992} and Rudin, Osher and Fatemi \cite{Rudin1992} are fundamental. 
We refer the reader to \cite{Buades2010} for a review and comparison of these methods.

% Among all these filters, the Neighborhood filter is the simplest, but yet useful, 
% method due to its compromise between denoising quality and computational speed. 
% Indeed, although it creates shocks and staircasing effects \cite{Buades2006a}, the computational cost 
% is by far lower than those of other integral kernel filters or PDE's based methods.
% 
% Since, usually, a single denoising step of the nonlocal filters is not enough, 
% an iteration  is performed according to several choices of 
% the iteration actualization, see \fer{def.GFF} and \fer{def.GFV} for two of such 
% strategies.

Nonlocal filters have been analyzed
from different points of view. For instance, Barash \cite{Barash2002}, Elad \cite{Elad2002},  Barash et al. \cite{Barash2004}, and Buades et al. \cite{Buades2006}
investigate the asymptotic relationship between the Yaroslavsky filter and the Perona-Malik PDE. Gilboa et al. \cite{Gilboa2008} study certain applications of 
nonlocal operators to image processing. In \cite{Peyre2008}, Peyr\'e establishes a relationship
between the non-iterative nonlocal filtering schemes and thresholding in adapted
orthogonal basis.
In a more recent paper, Singer et al. \cite{Singer2009}
interpret the Neighborhood filter as a stochastic diffusion process,
explaining in this way the attenuation of high frequencies in the processed images.

In \cite{gv2013} we heuristically introduced a denoising algorithm based in the Neighborhood filter but computed only on the level sets of the image, implying a large gain of computational effort. Later, in \cite{gv2014}, we reformulated this nonlocal filter in terms of the \emph{decreasing rearrangement} of the initial image, denoted by $u_*$,  which is defined as the inverse of the \emph{distribution function} $m_u(q) = \abs{\{\bx \in\O : u(\bx) >q\}}$, see Section 2 for the precise definition and some of its properties.

Realizing that the structure of level sets of $u$ is invariant through the Neighborhood filter operation as well as through the decreasing rearrangement of 
$u$ allowed us to rewrite \fer{def.NL}, for $w\equiv 1$, in terms of the one-dimensional integral expression
\begin{align*}
%\label{def.ubar}
 NF_h^* u(\bx)  = \frac{\int_{0}^{\abs{\O}}  \cK_h(u(\bx)-u_*(s)) u_*(s) ds}{\int_{0}^{\abs{\O}}  \cK_h(u(\bx)-u_*(s)) ds},
\end{align*}
which is computed jointly for all the pixels in each level set $\{\bx:u(\bx)=q\}$.

% Although from expression \fer{def.NF} is readily seen that only computation on level lines is needed to perform the filtering, the alternative expression in terms of the decreasing rearrangement offers room for further analysis of the iterative scheme.

Perhaps, the most important consequence of using the rearrangement was, apart from 
the large dimensional reduction, the reinterpretation of the Neighborhood filter as a 
\emph{local} algorithm. Thanks to this we proved, among others, the following properties for the most usual nonlinear iterative variant of the Neighborhood filter, see \fer{def.NL.i}:
\begin{itemize}
 \item The asymptotic behavior of the NF as a shock filter of the type introduced by \'Alvarez et al. \cite{Alvarez1994}, combined with a contrast loss effect.
 
 \item The contrast change character of the NF, i.e. the existence of a continuous and
 increasing function $g:\R\to\R$ such that $\NF^h (u(\bx))=g(u(\bx))$.
\end{itemize}

In this article we extend the use of rearranging techniques to other nonlocal filters.
Indeed, as noticed in \cite{gv2014}, even if the kernel $w$ is non-constant we may still use our approach by introducing the \emph{relative rearrangement} of the kernel with respect to the image, see Section 2 for definitions. 

In this way, we may express the general nonlocal filters embedded in formula \fer{def.NL}
in terms of one-dimensional integral expressions of the form 
\begin{align}
\label{def.ubar0}
 F_h^* u(\bx)  = \frac{\int_{0}^{\abs{\O}}  \cK_h(u(\bx)-u_*(s)) u_*(s) w(\bx,\cdot)_{*u} (s)ds}{\int_{0}^{\abs{\O}}  \cK_h(u(\bx)-u_*(s)) w(\bx,\cdot)_{*u} (s)ds}.
\end{align}
where $v_{*u}$ denotes the relative rearrangement of $v$ with respect to $u$. 

This sophisticated reformulation of the nonlocal filter in terms of one-dimensional integration may be of limited computational use since, in general, the filtering transformation requires computing for \emph{each} pixel, $\bx$,
the expensive term involving the relative rearrangement.

However, there are some particular filters, like the weighted Neighborhood filter, for which this reformulation offers a large gain of computational effort and gives a notable analtyical  insight into the filter functioning. In addition, in general, since the computation of 
\fer{def.ubar0} is based on the number of level lines (quantized levels) of $u$, when 
this number is small formula \fer{def.ubar0} may be more efficient than the direct implementation of \fer{def.NL}.

The plan of the article is the following. In Section~2 we introduce the notion 
of decreasing rearrangement and relative rearrangement and establish the equivalence between 
the usual pixel-based expression of the filter \fer{def.NL} and its rearranged formulation \fer{def.ubar0}

In Section~3, we provide a fully discrete algorithm to approximate by constant-wise functions the filter $F_h^* u(\bx)$ given by \fer{def.ubar0}, and thus the original equivalent filter $F_hu(\bx)$ given by \fer{def.NL}. We also prove the convergence of this discretization to the solution of the continuous setting.

In Section~4, we analyze the particular cases in which the kernel $w(\bx,\by)$ only 
depend on the integration variable $\by$, and thus may be considered as a weight function.
In this situation, we are able to extend most of the results proved for the Neighborhood filter
 in \cite{gv2014}, that is, for $w\equiv 1$. In particular, we show the asymptotic behavior of 
these filters as shock filters when $h\to 0$.

% Finally, let us emphasize that the reformulation of the nonlocal filters 
% we propose produces the same filtered image than that obtained through the classical 
% ones. However, there may be two important advantages in our approach:
% (i) the filtering process is performed through a one-dimensional
% integral, reducing largely the algorithm complexity, and (ii) the rearranged version exposes the nonlocal filter functioning to a deeper mathematical analysis.

\section{Nonlocal filters in terms of functional rearrangements}

\subsection{The decreasing rearrangement}

Let us denote by $\abs{E}$ the Lebesgue measure of any measurable set $E$.
For a Lebesgue measurable function $u:\O\to\R$, the function 
$q\in\R\to m_u(q) = \abs{\{\bx \in\O : u(\bx) >q\}}$ is called the \emph{distribution function} corresponding to $u$. 

Function $m_u$ is non-increasing and therefore admits a unique  generalized inverse, called the
\emph{decreasing rearrangement}. This inverse takes the usual pointwise meaning when 
the function $u$ has not flat regions, i.e. when $\abs{\{\bx \in\O : u(\bx) =q\}} =0$ for any $q\in\R$. In general, 
the decreasing rearrangement $u_*:[0,\abs{\O}]\to\R$ is given by:
\begin{equation*}
u_*(s) =\left\{
\begin{array}{ll}
 {\rm ess}\sup \{u(\bx): \bx \in \O \} & \qtext{if }s=0,\\
 \inf \{q \in \R : m_u(q) \leq s \}& \qtext{if } s\in (0,\abs{\O}),\\
 {\rm ess}\inf \{u(\bx): \bx \in \O \} & \qtext{if }s=\abs{\O}.
\end{array}\right.
 \end{equation*}
We shall also use the notation $\O_*=(0,\abs{\O})$.
Notice that since $u_*$ is non-increasing in $\bar\O_*$, it is continuous but at most a countable subset of  
$\bar\O_*$. In particular, it is right-continuous for all $t\in (0,\abs{\O}]$.

The notion of rearrangement of a function is classical and was introduced by Hardy, Littlewood and Polya \cite{Hardy1964}. Applications include the study of isoperimetric and variational inequalities \cite{Polya1951,Bandle1980,Mossino1984}, comparison of solutions of partial differential equations \cite{Talenti1976,Alvino1978,Vazquez1982,Diaz1992,Diaz1995,Alvino1996}, and others.
We refer the reader to the textbook \cite{Lieb2001} for the basic definitions.

Two of the most remarkable properties of the decreasing rearrangement are the  equi-measurability property, 
\begin{equation*}
%\label{prop.1}
 \int_\O F(u(\by))d\by = \int_0^{\abs{\O}} F(u_* (s))ds.
\end{equation*}
for any Borel function $F:\R\to\R_+$, and the contractivity
\begin{equation}
 \label{prop.2}
 \nor{u_*-v_*}_{L^p(\O_*)}\leq \nor{u-v}_{L^p(\O)},
\end{equation}
for $u,v\in L^p(\O)$, $p\in[1,\infty]$.

\subsection{Motivation}

Apart from the pure  mathematical interest, 
the reformulation of  nonlocal filters in terms of functional rearrangements 
is  useful for computational porpouses, specially when the level lines of $u$ are left invariant through the filter, i.e. when $u(\bx)=u(\by)$ implies  $F_h(u)(\bx)=F_h(u)(\by)$. Thus, in these cases, the filter is computed only for each (quantized) level line, instead of for each  pixel, meaning a large gaining of computational effort.

In the following lines, we provide a heuristic derivation of the nonlocal filter rearranged version, 
as first noticed in \cite{gv2014}. 
Under suitable assumptions, the coarea formula states 
\begin{equation*}
%\label{coarea}
 \int_\O g(\by)\abs{\nabla u(\by)} d\by =\int_{-\infty}^{\infty} \int_{u=t} g(\by) d\Gamma(\by) dt.
\end{equation*}
Taking $ g(\by)=\cK_h(u(\bx)-u(\by)) w (\bx,\by) u(\by)/\abs{\nabla u(\by)}$,
and using  $u(\bx) \in [0,Q]$ for all $\bx\in\O$ we get
\begin{align*}
 I(\bx)  :=\int_\O \cK_h(u(\bx)-u(\by)) w (\bx,\by) u(\by)d\by 
   =\int_{0}^{Q}  \cK_h(u(\bx)-t)  t 
  \int_{u=t}  \frac{w (\bx,\by)}{\abs{\nabla u(\by)}} d\Gamma(\by) dt.
\end{align*}
Introducing the change of variable $t=u_{*}(s)$  we find
\begin{align}
 I(\bx) & =-\int_{0}^{\abs{\O}}  \cK_h(u(\bx)-u_*(s))u_*(s) \frac{d u_* (s)}{ds}  \int_{u=u_*(s)}  \frac{w (\bx,\by)}{\abs{\nabla u(\by)}} d\Gamma(\by) ds \nonumber\\
 & =\int_{0}^{\abs{\O}}  \cK_h(u(\bx)-u_*(s)) u_*(s) w(\bx,\cdot)_{*u} (s)ds.
 \label{chco}
\end{align}
Here, the notation $v_{*u}$ stands for the \emph{relative rearrangement of $v$ with respect to $u$} which, 
under regularity conditions, may be expressed as 
\begin{equation}
\label{rrr}
 v_{*u}(s)  = \frac{\displaystyle\int_{u=u_*(s)}  \frac{v(\by)}{\abs{\nabla u(\by)}} d\Gamma(\by)}{ 
 \displaystyle\int_{u=u_*(s)}  \frac{1}{\abs{\nabla u(\by)}} d\Gamma(\by)},
\end{equation}
see the next section for details.
Transforming $C(\bx)$ in a similar way allows us to deduce 
\begin{align}
\label{def.NLR}
 \text{F}_{h} u(\bx)  = \frac{\int_{0}^{\abs{\O}}  \cK_h(u(\bx)-u_*(s)) u_*(s) w(\bx,\cdot)_{*u} (s)ds}{\int_{0}^{\abs{\O}}  \cK_h(u(\bx)-u_*(s)) w(\bx,\cdot)_{*u} (s)ds}.
\end{align}

\subsection{The relative rearrangement}

The relative rearrangement was introduced by Mossino and Temam \cite{Mossino1981} as the directional derivative of the decreasing rearrangement. Thus, if for $u\in L^1(\O)$ and $v\in L^p(\O)$, with $p\in [1,\infty]$, we consider the function $w:\O_*\to\R$ given by
\begin{equation*}
 w(s)=\int_{u>u_*(s)} v(\bx)d\bx + \int_0^{s-\abs{u>u_*(s)}} \big(v|_{u=u_*(s)}\big)_* (\sigma) d\sigma,
\end{equation*}
then the \emph{relative rearrangement of $v$ with respect to $u$}, $v_{*u}$, is defined as
\begin{equation*}
%\label{def.rr1}
 v_{*u}:=\frac{d}{ds} w \in L^p(\O_*).
\end{equation*}
This identity may be also understood as the weak $L^p(\O_*)$ directional derivative 
(weak* $L^\infty(\O_*)$, if $p=\infty$) 
\begin{equation}
\label{def.rr}
 v_{*u}= \lim_{t\to0}\frac{(u+tv)_* - u*}{t}. 
\end{equation}
Under the additional assumptions $u\in W^{1,1}(\O)$ and $\abs{\{\by\in\O :\nabla u(\by)=0 \}}=0$, i.e. the non-existence of flat regions of $u$, the identity \fer{rrr} is well defined. In this case the relative rearrangement 
represents an averaging procedure of the values of $v$ on the level lines of $u$ labeled by 
the superlevel sets measure, $s$.

When formula \fer{rrr} does not apply, that is in flat regions of $u$, we may resort to 
identity \fer{def.rr} to interpret the relative rearrangement in flat regions as the 
decreasing rearrangement of $v$ restricted to such sets.

After the seminal work of Mossino and Temam \cite{Mossino1981}, 
the relative rearrangement was further studied by Mossino and Rakotoson \cite{Mossino1986}
and applied to several types of problems, among which those related to variable exponent spaces and functional properties, see Rakotoson et al. \cite{Fiorenza2007,Fiorenza2009,Rakotoson2010,Rakotoson2012}, or nonlocal formulations of plasma physics problems related to nuclear fusion devices, see D\'{\i}az et al.\cite{Diaz1996,Diaz1998,Diaz2004}.

% {\redtext To our best knowledge, in this article we introduce the use of the relative rearrangement for the analysis of image nonlocal filters for the first time, apart from the note already appearing in \cite{gv2014}.}

In the rest of the article, we shall make an extensive use of results appearing in the monograph on the relative rearrangement by Rakotoson \cite{Rakotoson2008}.

\subsection{A general result}
The main assumption we implicitly made for the heuristic deduction of formula \fer{def.NLR} is 
the condition $\abs{\{\by\in\O :\nabla u(\by)=0 \}}=0$, i.e. the non-existence of flat regions of $u$, 
which gives sense on one hand to formula \fer{rrr}, and on the other hand,  
allow us to obtain the strictly decreasing behaviour of $u_*$ which justifies the change of variable in \fer{chco}.

Our first result is that formulas \fer{def.NL} and \fer{def.NLR} are equivalent under  weaker hypothesis. Due to the nature of our application, we keep the assumption on the boundedness of
$u$ and $w$ in $L^\infty$, although these can be also weakened to less regular $L^p$ spaces.

\begin{theorem}
\label{th.equivalence}
 Let $\O\subset\R^d$ be an open and bounded set, $d\geq 1$, $\cK\in L^\infty(\R,\R_+)$ 
 and $w\in L^\infty(\O\times\O,\R_+)$. Assume that $u\in L^\infty(\O)$ is, without loss of generality, non-negative. Consider $F_hu(\bx)$ given by \fer{def.NL} and 
 \begin{align}
\label{def.ubar}
 F_h^* u(\bx)  = \frac{\int_{0}^{\abs{\O}}  \cK_h(u(\bx)-u_*(s)) u_*(s) w(\bx,\cdot)_{*u} (s)ds}{\int_{0}^{\abs{\O}}  \cK_h(u(\bx)-u_*(s)) w(\bx,\cdot)_{*u} (s)ds}.
\end{align}
Then $F_h^* u(\bx) = F_h u(\bx)$ for a.e. $\bx \in \O$.
\end{theorem}
\no\emph{Proof. } Let $f\in L^\infty(\R)$ and $b\in L^\infty(\O)$.
We start showing
\begin{equation}
\label{expr.0}
 \int_\O f(u(\by))b(\by)d\by = \int_0^{\abs{\O}} f(u_*(s))b_{*u}(s)ds.
\end{equation}
Consider the sets of flat regions of $u$ and $u_*$, 
\begin{equation}
 \label{def.P}
 P=\bigcup_{i\in D} P_i,\quad P_i=\{\by\in\O : u(\by)=q_i\},
\end{equation}
and $P_*=\cup_{i\in D} P^*_i$, with , $P^*_i=\{s\in\O_* : u_*(s)=q_i\}$, where the subindices set $D$ is, at most, countable.
According to \cite[Lemma 2.5.2]{Rakotoson2008}, we have 
\begin{align}
\label{expr.1}
 \int_0^{\abs{\O}} f(u_*(s))b_{*u}(s)= \int_{\O\backslash P} f(u_* (m_u(u(\by))))b(\by)d \by+ 
 \sum_{i\in D}\int_{P_i} M_{b_i}(h_i)(\by)b(\by)d\by,
\end{align}
where $b_i=b|_{P_i}$, $h_i(s)=f(u_*(s_i'+s))$ for $s\in [s_i',s_i''):=P_i^*$,  and
\begin{equation*}
M_{b_i}(h_i)(\by)=\left\{
\begin{array}{ll}
 h_i(m_{b_i}(b_i(\by))) & \text{if } \by\in P_i\backslash Q^i,\\
 \displaystyle\frac{1}{\abs{Q_j}}\int_{\sigma_j'}^{\sigma_j''} h_i(s)ds, & \text{if } \by\in Q^i_j,\\
\end{array}
\right.
\end{equation*}
where $Q^i=\cup_{j\in D'}Q_j^i$, with $Q_j^i$ the flat regions of $b_i$ and $ [\sigma_j',\sigma_j''):=Q_i^{j*}$. In \fer{expr.1},  since the functions $u_*$ and 
$m_u$ are strictly decreasing and inverse of each other in the set $\O\backslash P$, we obtain
\begin{align*}
%\label{expr.1}
 \int_{\O\backslash P} f(u_* (m_u(u(\by))))b(\by)d \by=
 \int_{\O\backslash P} f(u(\by))b(\by)d \by.
\end{align*}
In the flat regions of $u$ and $u_*$ we have, on one hand,
\begin{align*}
 \int_{P} f(u(\by))b(\by)d \by= \sum_{i\in D}\int_{P_i} f(u(\by))b(\by)d \by=
 \sum_{i\in D}f(q_i)\int_{P_i} b(\by)d \by.
\end{align*}
And, on the other hand, since $h_i(s)=f(u_*(s_i'+s))=f(q_i)$ for $s\in P_i^*$, 
\begin{align*}
 \sum_{i\in D}\int_{P_i} M_{b_i}(h_i)(\by)b(\by)d\by= & \sum_{i\in D}  \left(\int_{P_i\backslash Q^i}   h_i(m_{b_i}(b_i(\by))) b(\by) d\by + \right. 
 \left.\sum_{j\in D'} \frac{1}{\abs{Q_j}}\int_{\sigma_j'}^{\sigma_j''} h_i(s)ds \int_{ Q_j^i} b(\by)d\by\right) \\
 =  & \sum_{i\in D}   \left(f(q_i) \int_{P\backslash Q^i}  b(\by) d\by+\right. 
 \left. f(q_i) \sum_{j\in D'} \int_{Q_j^i} b(\by)d\by\right) \\
 = & \sum_{i\in D}f(q_i)\int_{P_i} b(\by)d \by.
\end{align*}
Therefore, both sides of \fer{expr.0} are equal. 

Finally, for fixed $\bx\in \O$ set $b(\by) = w(\bx,\by)$ and first, 
$f(t)=\cK_h(u(\bx)-t)t$, for $t\geq0$  to obtain, using the identity \fer{expr.0}, the equality between the numerators of \fer{def.NL} and \fer{def.ubar}, and second,  $f(t)=K_h(u(\bx)-t)$ to obtain the equality between the denominators of those expressions.
$\Box$

\begin{remark}
 As deduced in the proof, identity \fer{expr.0} follows from  \cite[Lemma 2.5.2]{Rakotoson2008}. In fact, a little more than \fer{expr.0} may be obtained.
 Let $f,b\in L^\infty(\O)$. Then, if $f$ is constant in the flat regions of $u$, that is
 $f|_{P_i}=f_i=const$, then 
\begin{equation}
\label{expr.10}
 \int_0^{\abs{\O}} f(s)b_{*u}(s)ds=\int_\O f(m_u(u(\by)))b(\by)d\by+\sum_{i\in D}f(q_i)\int_{P_i} b(\by)d \by.
\end{equation}
\end{remark}

\section{Constant-wise discretization and convergence}

In this section we provide a fully discrete algorithm to approximate the filter 
$F_h^*$ given by \fer{def.ubar}, and thus the original equivalent filter $F_h$ given by \fer{def.NL}, as proved in Theorem~\ref{th.equivalence}.

In Theorem~\ref{th.convergence} we prove that if the initial image $u$ has a finite number of flat regions, that is if the set $D$ given in \fer{def.P} is finite, then we may approximate $u$ and $w$ by constant-wise functions $u_n,~w_{m}$ which have a finite number of levels and such that 
\begin{equation*}
%\label{th.conv3}
 F^*_{h,m} u_n (\bx)\to F_h^*u (\bx)
\end{equation*}
where $F^*_{h,m}$ is the discrete version of $F^*_h$, see \fer{def.fhm}.

This result gives sense to Theorem~\ref{th.approximation}, in which we produce a finite discrete formula for the approximation of $F^*_h u(\bx)$, and thus of $F_hu(\bx)$. This formula is what can actually be used for the numerical experimentation.

\begin{theorem}
\label{th.convergence}
Let $u,w\in L^\infty(\O)$ be nonnegative and assume that $u$ has a finite number of flat regions. Then, there exist sequences of constant-wise functions $u_n,~w_m$, with a finite number of flat regions, such that $u_n\to u$ strongly in $L^\infty(\O)$, $w_m\to w$ strongly in $L^\infty(\O\times\O)$,  and 
\begin{equation}
\label{th.conv3}
 F^*_{h,m} u_n \to F_h^*u \qtext{a.e. in }\O\qtext{and strongly in } L^\infty(\O),
\end{equation}
where
\begin{align}
\label{def.fhm}
 F_{h,m}^* u(\bx)  = \frac{\int_{0}^{\abs{\O}}  \cK_h(u(\bx)-u_*(s)) u_*(s) w_m(\bx,\cdot)_{*u} (s)ds}{\int_{0}^{\abs{\O}}  \cK_h(u(\bx)-u_*(s)) w_m(\bx,\cdot)_{*u} (s)ds}.
\end{align}
\end{theorem}
\no\emph{Proof. }We split the proof in several steps.

\no\emph{Step 1. }We use the construction of the sequence of constant-wise functions $u_n$ given in \cite[Th. 7.2.1]{Rakotoson2008}. In our case, the construction is simpler because $u$ has a finite number of flat regions, implying that $u_n$ has a finite number of levels.

In any case, this construction is such that $u_n\to u$ a.e. in $\O$ and strongly in $L^\infty(\O)$, and $w_{*u_n}\to w_{*u}$ weakly* in $L^\infty(\O_*)$. Besides, 
due to the strong continuity of the decreasing rearrangement \fer{prop.2}, we also have 
$(u_n)_*\to u_*$ strongly in $L^\infty(\O_*)$. Therefore, we readily see first that 
\begin{equation*}
 F_h^*u_n(\bx) \to F_h^* u(\bx) \qtext{for a.e. }\bx\in \O,
\end{equation*}
and then, due to the dominated convergence theorem,
\begin{equation}
\label{conv.1}
 F_h^*u_n \to F_h^* u \qtext{strongly in } L^\infty(\O).
\end{equation}

\no\emph{Step 2. }We consider a sequence of constant-wise functions with a finite number of levels, $w_m$, such that $w_m\to w$ strongly in $L^\infty(\O\times\O)$. Due to the contractivity property of the relative rearrangement, see \cite{Mossino1981}, we also have, for a.e. $\bx\in\O$, $w_m(\bx,\cdot)_{*v}\to w(\bx,\cdot)_{*v}$ strongly in $L^\infty(\O)$, for any $v\in L^\infty(\O)$. Thus, as $m\to \infty$,
\begin{align*}
%\label{def.ubar}
 F_{h,m}^* u_n(\bx)  =F_{h}^* u_n(\bx) \qtext{for a.e. }\bx\in \O,
\end{align*}
and, again, the dominated convergence theorem implies
\begin{equation}
\label{conv.2}
 F_{h,m}^*u_n \to F_h^* u_n \qtext{strongly in } L^\infty(\O).
\end{equation}

\no\emph{Step 3. }In view of \fer{conv.1} and \fer{conv.2}, we have
\begin{align*}
 \abs{F_{h,m}^*u_n -F_h^*u}\leq \abs{F_{h,m}^*u_n -F_h^*u_n}+ \abs{F_{h}^*u_n -F_h^*u}\to0
\end{align*}
as $m\to\infty$ and $n\to\infty$, so \fer{th.conv3} follows.
$\Box$
\begin{remark}
 Theorem~\ref{th.convergence} may be extended to the case in which $u$ has a countable 
 number of flat regions. However, the construction in \cite{Rakotoson2008} then implies
 that each element of the sequence of constant-wise functions $u_n$ has also a countable number of levels. Since our aim is providing a \emph{finite} discretization for numerical implementation,   such a sequence is not appropriate. 
\end{remark}

In the following theorem we produce a discrete numerical formula for computing 
the nonlocal filter for each pair $(u_n,w_m)$ of the sequences given in Theorem~\ref{th.convergence}. 

The main difficulty of computing formula \fer{def.fhm} the determination of 
the relative rearrangement. However, in the case of constant-wise functions with 
a finite number of levels this computation is simplified thanks to identity \fer{def.rr}, 
which may be easily applied to this situation, as shown in \cite[Th.7.3.4]{Rakotoson2008}

In few words, for the case of constant-wise functions $u$ and $v$, 
the relative rearrangement $v_{*u}$ may be computed as the decreasing rearrangement of $v$ 
restricted to the level sets of $u$.

%We drop the subindices $n$ and $m$ from these sequences for clarity.

\begin{theorem}
\label{th.approximation}
Let $u\in L^\infty(\O)$ be a constant-wise function quantized in $n$ levels labeled by $q_i$, with $\max (u)=q_1 >\ldots > q_n=0$. That is $u(\bx)=\sum_{i=1}^n q_i\chi_{E_i} (\bx),$
% \begin{equation*}
%  u(\bx)=\sum_{i=1}^n q_i\chi_{E_i} (\bx),
% \end{equation*}
where $E_i$ are the level sets of $u$, 
 \begin{equation*}
 %\label{def.levelsets}
E_i=\{\bx \in \O : u(\bx)=q_i \},\quad i=1,\ldots,n.
\end{equation*}
Similarly, let $w\in L^\infty(\O\times\O)$ be constant-wise and 
quantized in $m$ levels $r_j$, with $\max (w)=r_1 >\ldots > r_m=\min(w)\geq0$.
For each $\bx\in \O$, consider  the partition of $E_i$ given by $F_j^i(\bx)=\{\by \in E_i : w(\bx,\by)=r_j \}.$
% \begin{equation}
%  %\label{def.levelsets}
% F_j^i(\bx)=\{\by \in E_i : v(\by)=r_j(\bx) \}.
% \end{equation}
Then, for each $\bx\in E_k$, $k=1,\ldots,n$
\begin{align}
\label{def.NLRD}
 F^*_h u(\bx)=\frac{\sum_{i=1}^n \sum_{j= 1}^m \cK_h(q_k-q_i) q_i r_j \abs{F_j^i(\bx)}}{\sum_{i=1}^n \sum_{j= 1}^m \cK_h(q_k-q_i)  r_j \abs{F_j^i(\bx)}}.
\end{align}
\end{theorem}
\no\emph{Proof. }Since $u$ is constant-wise, the decreasing rearrangement of $u$ is 
constant-wise too, and given by 
\begin{equation*}
 u_*(s)=\sum_{i=1}^n q_i\chi_{I_i}(s), 
\end{equation*}
with $I_i=[a_{i-1},a_i)$ for $i=1,\ldots,n$, and $a_0=0$, $a_1=\abs{E_1}$, $a_2=\abs{E_1}+\abs{E_2}$,$\ldots$,$a_n=\sum_{i=1}^n \abs{E_i}=\abs{\O}$. 
It is covenient to introduce here the cumulative sum of sets measures 
\begin{equation*}
 \cum(E_\circ,0)= 0,\qtext{and}\quad \cum(E_\circ,i)=\sum_{k=1}^i \abs{E_k}, \quad i=1,...,n,
\end{equation*}
where the symbol $\circ$ denotes the summation variable. Thus, $a_i=\cum(E_\circ,i)$.

% As above mentioned, each $u$-level set, $E_i$, may be partitioned based on the $v$-level sets  as $E_i=\cup_{j=1}^m F_j^i(\bx)$, 
% % with 
% % \begin{equation}
% %  %\label{def.levelsets}
% % F_j^i=\{\by \in E_i : v(\by)=r_j \}.
% % \end{equation}
% and then, if $\by\in F_j^i(\bx)$ we have $u(\by)=q_i$ and $v(\by)=r_j(\bx)$. 

We consider, for fixed $\bx\in \O$ and $t>0$, the funtion $H(\by):=u(\by)+tw(\bx,\by)$. Since both $u$ and $w$  are constant-wise with a finite number of levels, we have, for $t$ small enough
\begin{equation*}
 q_{i+1}< q_i+tr_j< q_{i-1},
\end{equation*}
implying that each level set of $H$ is included in one and only one level set of $u$.
Thus, we have the orderings
\begin{itemize}
 \item For each $j$, if $i_1>i_2$   and $\by\in F_j^{i_1}(\bx),~ \bar \by \in F_j^{i_2}(\bx)$
 then  $H(\by)<H(\bar \by)$.
 
 \item For each $i$, if $j_1>j_2$   and $\by\in F_{j_1}^{i}(\bx),~ \bar \by \in F_{j_2}^{i}(\bx)$
 then  $H(\by)<H(\bar \by)$.
\end{itemize}
With these observations we may compute the decreasing rearrangement of $H$ as follows.
For instance, for $s\in I_1=[0,\abs{E_1})$ we have (omitting $\bx$ from the notation $F_i^j(\bx)$)
\begin{align*}
 H_*(s)=\left\{
 \begin{array}{ll}
  q_1+tr_1 & \text{if } s\in [0,\abs{F_1^1})=[\cum(F^1_\circ,0),\cum(F^1_\circ,1)),\\
  q_1+tr_2 & \text{if } s\in [\abs{F_1^1},\abs{F_1^1}+\abs{F_2^1})=
  [\cum(F^1_\circ,1),\cum(F^1_\circ,2)),\\
  \ldots & \ldots \\
  q_1+tr_m & \text{if } s\in [\cum(F^1_\circ,m-1),\cum(F^1_\circ,m))=[\cum(F^1_\circ,m-1),\abs{E_1}).
 \end{array}
\right.
\end{align*}
In general, we may write for $i=1,\ldots,n$ and $j=1,\ldots,m$,
\begin{align*}
 H_*(s)= q_i+tr_j \qtext{if } s\in J_j^i(\bx),
\end{align*}
where $J_j^i(\bx):=\left[b_{j-1}^i(\bx),b_j^i(\bx)\right)$,
with $b_j^i(\bx):=a_{i-1}+\cum(F_\circ^i(\bx),j)$.
Observe that $b_m^i(\bx)=a_i$. 
Finally, since $J_j^i(\bx)\subset E_i$ we have for $s\in J_j^i(\bx)$
\begin{align*}
 \frac{H_*(s)-u_*(s)}{t}=r_j,\qtext{implying}\quad w(\bx,\cdot)_{*u}(s) = r_j.
\end{align*}
We are now in disposition to compute formula \fer{def.ubar}. For $\bx \in E_k$,
\begin{align*}
 \int_{0}^{\abs{\O}}  \cK_h(u(\bx)-u_*(s)) u_*(s) w(\bx,\cdot)_{*u} (s)ds = \sum_{i=1}^n \sum_{j= 1}^m \cK_h(q_k-q_i) q_i r_j \abs{J_j^i(\bx)}
\end{align*}
In a similar way we obtain 
\begin{align*}
 C(\bx)=\int_{0}^{\abs{\O}}  \cK_h(u(\bx)-u_*(s)) w(\bx,\cdot)_{*u} (s)ds = \sum_{i=1}^n \sum_{j= 1}^m \cK_h(q_k-q_i)  r_j \abs{J_j^i(\bx)}
\end{align*}
and therefore, using the definition of the sets $J_j^i(\bx)$ we obtain, \fer{def.NLRD}.
$\Box$

\subsection{Examples}

As it clear from formula \fer{def.NLRD}, the main difficulty for its computation is the determination of the measures of the sets $F_{j}^i(\bx)$, which must be computed for each $\bx\in \O$. 

The formula also provides the complexity of the algorithm. If $N$ is the number of pixels of the image, then the complexity is of the order $O(Nnm)$, where $n$ is the number of levels of the image and $m$ is the number of levels of the kernel.
Let us examine some examples.

\bigskip

\no\emph{The Neighborhood filter. }In this case, $w(\bx,\by)\equiv 1$,   and therefore
$j=1$ and $F_j^i(\bx)=E_i$ is independent of $\bx$ for all $i=1,\ldots,n$. Thus, formula \fer{def.NLRD} is computed only on the level sets of $u$, that is, for all $\bx\in E_k$
\begin{align*}
%\label{def.NLRD}
 F^*_h u(\bx)=\frac{\sum_{i=1}^n  \cK_h(q_k-q_i) q_i  \abs{E_i}}{\sum_{i=1}^n  \cK_h(q_k-q_i)  \abs{E_i}}
\end{align*}
In this case, the complexity is of order $O(n^2)$.

\bigskip

\no\emph{The weighted Neighborhood filter. }
Here, $w(\bx,\by)\equiv \bar w(\by)$, and therefore
$F_j^i(\bx)$ is independent of $\bx$ for all $i=1,\ldots,n$, $j=1,\ldots,m$. 
Thus, formula \fer{def.NLRD} is computed again only on the level sets of $u$, that is, for all $\bx\in E_k$
\begin{align*}
%\label{def.NLRD}
 F^*_h u(\bx)=\frac{\sum_{i=1}^n \sum_{j= 1}^m \cK_h(q_k-q_i) q_i r_j \abs{F_j^i}}{\sum_{i=1}^n \sum_{j= 1}^m \cK_h(q_k-q_i)  r_j \abs{F_j^i}}
\end{align*}
The complexity is of order $O(n^2m)$.

\bigskip

\no\emph{The Yaroslavsky filter. }
In this case, $w(\bx,\by)= \chi_{B_\rho(\bx)}(\by)$, and therefore
there are only two levels $r_1=1,~r_2=0$ of $w$ corresponding to the sets 
\begin{equation*}
 %\label{def.levelsets}
F_1^i(\bx)=\{\by \in E_i : \abs{\bx-\by)}<\rho \},\quad 
F_2^i(\bx)=\{\by \in E_i : \abs{\bx-\by)}\geq\rho \},
\end{equation*}
Thus, formula \fer{def.NLRD} reduces to: for each $\bx\in E_k$, $k=1,\ldots,n$
\begin{align*}
%\label{def.NLRD}
 F^*_h u(\bx)=\frac{\sum_{i=1}^n  \cK_h(q_k-q_i) q_i  \abs{F_1^i(\bx)}}{\sum_{i=1}^n  \cK_h(q_k-q_i)   \abs{F_1^i(\bx)}}
\end{align*}
The complexity is of order $O(Nn)$.

\bigskip

\no\emph{The Bilateral filter. }In this case, $w(\bx,\by)= \exp({\abs{\bx-\by}^2/\rho^2})$ and therefore there is a continuous range of levels for $w$. However, for computational porpouses the range of $w$ is quantized to some finite
number of levels, determined by the size of $\rho$. Thus, the full formula \fer{def.NLRD}
must be used in this case. The resulting complexity is of  order $O(Nnm)$.

\section{The Neighborhood and weighted Neighborhood filters}

  The next results are particularized to the cases in which the introduction of the rearranged formulation \fer{def.NLR} implies an important gain in the algorithmic complexity. This happens when the function $w$ is a weight function 
  instead of window functions, i.e. if $w(\bx,\by)\equiv w(\by)$. 
%As shown in \cite{gv2014}, a very detailed behavior of the nonlocal filter may be deduced. 

This simpification is the case of, for instance, the Neighborhood filter ($w\equiv 1$) or the weighted Neighborhood filter (non-negative $w \in L^\infty(\O)$). In these cases, since one application of the filter is usually not enough, the following iterative scheme is considered. Set $u_0=u$, the initial image. For $n\in \N$,
\begin{align}
 \label{def.NL.i}
    u_{n+1}(\bx)= 
   \frac{1}{C_n(\bx)}\int_\O \cK_h(u_n(\bx)-u_n(\by)) u_n(\by) w (\by) d\by,
 \end{align}
with $C_n(\bx)= \int_\O \cK_h(u_n(\bx)-u_n(\by)) w (\by) d\by$. 

Straightforward adaptations of the proofs of Theorems 1 and 2 and Corollaries 1 and 2  of \cite{gv2014}, proved for $w\equiv 1$, 
imply similar results for the general case of  a
weight functions $w\geq 0$. We list here the most salient properties stated in these 
results, to which we shall refer as to \textbf{Properties (P)}.
We use the following notation for the level sets of $u$ given in terms of the levels of $u_*$:
\begin{equation*}
 %\label{def.levelsets}
L_t(u)=\{\by \in \O : u(\by)=u_*(t) \},\qtext{for }t \in \bar \O_*.
\end{equation*}

\begin{enumerate}
 \item  The iterative scheme \fer{def.NL.i} may be computed only on the level sets of $u$ as follows: if $\bx \in L_t(u)$ for some $t\in [0,\abs{\O}]$, we set $u_{n+1}(\bx)=   v_{n+1}(t)$, with
 \begin{equation}
\label{def.NFstar}
  v_{n+1}(t) = \frac{1}{c_n(t)} \int_0^{\abs{\O}} \cK_h(v_{n}(t)-v_{n}(s))v_{n}(s)w_{*u} (s)ds,
\end{equation}
  $c_n(t)=\int_0^{\abs{\O}} \cK_h(v_{n}(t)-v_{n}(s))w_{*u} (s)ds$,
  and $v_0 =u_*$.
  
  \item Under suitable assumptions on the kernel $\cK$, in which the Gaussian kernel is included, if $v_0\in W^{1,p}(0,\abs{\O})$ for some $p\geq 1$ then 
 \begin{enumerate}
  \item[(i)] $v_{n+1}\in W^{1,p}(0,\abs{\O})$,  and if $v'_0(t)=0$ then $v'_{n+1}(t)=0$.
  
    \item[(ii)]  $v'_{n+1} \leq 0$ a.e. in $(0,\abs{\O})$, and if $v'_0(t)<0$ then $v'_{n+1}(t)<0$.
    
    \item[(iii)] If $v_0\in C^m([0,\abs{\O}])$ and $\cK\in C^m(\R)$ then 
    $v_{n+1}\in C^m([0,\abs{\O}])$ for all $n$.
    
    \item[(iv)] For each $n\in\N$, there exists a strictly increasing function $g:\R\to\R$, a \emph{contrast change}, such that $u_{n+1}(\bx)=g(u(\bx))$, where $u_{n+1}$ is given by \fer{def.NL.i}.
 \end{enumerate}
  
\end{enumerate}

In the following theorem we establish a correspondence between the nonlocal diffusion 
scheme \fer{def.NFstar} and local diffusion, establishing an asymptotic behavior of the 
filter, when $h\to0$ as a shock filter. 

Although more general assumptions on $\cK$ may be prescribed, see Remark~\ref{remth.pde}, we estate the following result for the Gaussian kernel, for clarity. We also ask for further regularity on $u_*$ and $w_{*u}$.

\begin{theorem}
\label{th.pde}
Let $v_0=u_* \in C^3(\bar\O_*)$ and $w_{*u}\in C^2(\bar\O_*)$ be such that $v_0'<0$ and $w_{*u}>0$ in $[0,\abs{\O}]$. Set  $ \cK(\xi)=\text{e}^{-\xi^2}$.
Then, for all $t\in \O_*$, there exist positive constants $\alpha_1$, and $\alpha_2$, independent of $h$ such that
\begin{align}
\label{app.NFstar}
v_{n+1}(t) = v_{n}(t)&+\alpha_1 \frac{\tilde k_h(t)v_{n} ' (t)}{w_{*u}(t)} \big( h+O(h^{3/2}) \big) - \alpha_2 \frac{v_{n} '' (t)}{(v_{n} ' (t))^2} h^2\\
&  + \alpha_2 \frac{w_{*u}'(t)}{w_{*u}(t)v_n'(t)}h^2 + O(h^{5/2}), \nonumber
\end{align}
with 
\begin{equation}
\label{def.ktilde}
 \tilde k_h(t)= \frac{w_{*u}(\abs{\O})\cK_h(v_n(t)-v_n(\abs{\O}))}{v_n'(\abs{\O})}-
 \frac{w_{*u}(0)\cK_h(v_n(t)-v_n(0))}{v_n'(0)},
\end{equation}
and with $\alpha_1\approx 1/\sqrt{\pi}$, and $\alpha_2\approx 1$.
\end{theorem}

 For $w\equiv 1$, and thus $w_{*u}\equiv 1$, this result was proven in \cite{gv2014}. The second and third terms at the right hand side of \fer{app.NFstar} were interpreted, respectively,  as a (border) loss of contrast, and an  anti-diffusive shock filter term similar to that introduced by Alvarez and Mazorra \cite{Alvarez1994}. 
 
 However, for a general weights $w$, the fourth term is more difficult to interpret due specially to the unclear meaning of the derivative $w_{*u}'$. Just to gain some insight, let us assume that $w(\bx)=f(u(\bx))$ for some contrast change function, $f$. Then, $w_{*u}=f(u_*)$. 
 By point 2(iii) of Properties (P), for each step $n$ there exists another contrast change, $g$, such that $v_n=g(u_*)$. Then we have
 \begin{align*}
  \frac{w_{*u}'(t)}{w_{*u}(t)v_n'(t)}=\frac{f'(u_*(t))}{f(u_*(t))g'(u_*(t))},
 \end{align*}
which corresponds to a nonnegative source term.

% Let us finally mention that the condition $w_{*u}>0$ can be weakened in some cases. For instance, when $w_{*u}(0)=w_{*u}(\abs{\O})=0$ and $\ln w_{*u} \in C^1([0,\abs{\O}])$.

\emph{Proof of Theorem~\ref{th.pde}. }
We may rewrite the iterative scheme  \fer{def.NFstar} as
\begin{align}
\label{th2.1}
 v_{n+1}(t)-v_n(t)= %\\ 
  \frac{1}{c_n(t)}\int_0^{\abs{\O}} \cK_h(v_n(t)-v_n(s))(v_n(s)-v_n(t)) w_{*u} (s)ds.%\nonumber
\end{align}
Due to (P) we have $v_{n}'<0$ in $(0,\abs{\O})$, and $v_n(0,\abs{\O})\subset v_0(0,\abs{\O})$.
Let us denote the  inverse of $v_{n}$ by $v_n^{-1}$.
Using the change of variable $s=v_n^{-1}(q)$ and writing
$t=v_n^{-1}(z)$, we obtain from \fer{th2.1}
\begin{equation}
\label{th2.3}
 v_{n+1}(t)-v_n(t)=\frac{I_1(z)}{I_2(z)},
\end{equation}
with 
\begin{align*}
 & I_1(z)= \int_{v_n(\abs{\O})}^{v_n(0)}  \cK_h(z-q)(q-z)\frac{w_{*u} (v_n^{-1}(q))}{v_{n} ' (v_n^{-1}(q))}dq,\\ 
 & I_2(z)=\int_{v_n(\abs{\O})}^{v_n(0)} \cK_h(z-q)\frac{w_{*u} (v_n^{-1}(q))}{v_{n} ' (v_n^{-1}(q))}dq.
\end{align*}
Using the Gaussian explicit form of $\cK$ and integrating by parts, we obtain 
\begin{align}
\label{th2.i1}
 I_1(z)= \frac{h^2}{2}\big( \tilde k_h (v_n^{-1}(z)) 
  +\int_{v_n(\abs{\O})}^{v_n(0)} \cK_h(z-q)f(q)dq\big),
\end{align}
with $\tilde k_h$ given by \fer{def.ktilde}, and
\begin{equation*}
 f(q)=\frac{w_{*u}' (v_n^{-1}(q))}{(v_{n} ' (v_n^{-1}(q)))^2}-\frac{w_{*u}(v_n^{-1}(q))v_{n} '' (v_n^{-1}(q))}{(v_{n} ' (v_n^{-1}(q)))^3}.
\end{equation*}
Let us also introduce
\begin{equation*}
 g(q)=\frac{w_{*u}(v_n^{-1}(q))}{v_{n} ' (v_n^{-1}(q))}.
\end{equation*}
By assumption, $f$ and $g$ 
are bounded in $[v_n(\abs{\O}),v_n(0)]$ and by (P) they are also continuously differentiable in $(v_n(\abs{\O}),v_n(0))$.

Consider the interval $J_h=\{q: \abs{z-q}<\sqrt{h}\}$.
By well known properties of the Gaussian kernel, we have
\begin{equation}
 \label{gauss.1}
 \kappa(h): = \int_{J_h} \cK_h(z-q) dq <  \int_\R \cK_h(q) dq = h\sqrt{\pi},
\end{equation}
and
\begin{equation}
 \label{gauss.2}
\cK_h(z-q)\leq  \text{e}^{-1/h} \quad\text{if}\quad  q\in J_h^C=\{q:\abs{z-q}\geq\sqrt{h}\}.
\end{equation}
 In particular, from \fer{gauss.2} we get 
\begin{equation}
\label{th2.4}
 \left| \int_{J_h^C} \cK_h(z-q)f(q)dq \right| < O(h^\alpha)\qtext{for any }\alpha >0.
\end{equation}
Taylor's formula implies 
\begin{align*}
  \int_{v_n(\abs{\O})}^{v_n(0)}   \cK_h (z-q)  f(q)dq  
  =  \int_{J_h} \cK_h(z-q) (f(z) +O(\sqrt{h})) dq 
  +  \int_{J_h^C} \cK_h(z-q)f(q)dq .
\end{align*}
Therefore, from \fer{th2.i1}, \fer{gauss.1} and \fer{th2.4} we deduce
\begin{align*}
I_1(z)=\frac{h^2}{2} \Big( \tilde k(v_n^{-1}(z)) + f(v_n^{-1}(z)) \kappa(h) + O(h^{3/2})\Big).
\end{align*}
Similarly,
\begin{align*}
 I_2(z) = & \int_{v_n(\abs{\O})}^{v_n(0)}  \cK_h(z-q)g(q)dq =
 \int_{J_h} \cK_h(z-q) (g(z)+O(\sqrt{h})) dq 
  + \int_{J_h^C} \cK_h(z-q)g(q)dq  \\
  = & g(v_n^{-1}(z)) \kappa(h) + O(h^{3/2}).
\end{align*}
Then, 
the result follows from \fer{th2.3} substituting $z$ by $v_n(t)$. $\Box$

\begin{remark}
 \label{remth.pde}
 Theorem~\ref{th.pde} may be extended to Lipschitz continuous decaying kernels satisfying the growth condition 
 \begin{equation*}
  %\label{ext.kernel}
   \cK(s)\leq \frac{k_0}{1+\abs{s}^p},\qtext{for some }p>1.
\end{equation*}
See \cite{gv2014} for details.
\end{remark}


\begin{thebibliography}{25}
\providecommand{\natexlab}[1]{#1}
\providecommand{\url}[1]{{#1}}
\providecommand{\urlprefix}{URL }
\expandafter\ifx\csname urlstyle\endcsname\relax
  \providecommand{\doi}[1]{DOI~\discretionary{}{}{}#1}\else
  \providecommand{\doi}{DOI~\discretionary{}{}{}\begingroup
  \urlstyle{rm}\Url}\fi
\providecommand{\eprint}[2][]{\url{#2}}

\bibitem{Alvarez1994}
\'Alvarez L, Mazorra L (1994) Signal and image restoration using shock filters
  and anisotropic diffusion. Siam J Numer Anal 31(2):590--605


\bibitem{Alvarez1992}
\'Alvarez L, Lions PL, Morel JM (1992) Image selective smoothing and edge
  detection by nonlinear diffusion. ii. Siam J Numer Anal 29(3):845--866

  \bibitem{Alvino1978}
  Alvino A, Trombetti, G (1978)
Sulle migliori costanti di maggiorazione per una classe di equationi ellittiche
degeneri.
Ricerche Mat 27:413--428

\bibitem{Alvino1996}
Alvino A, D\'{\i}ıaz JI, Lions PL, Trombetti, G (1996)
Elliptic Equations and Steiner Symmetrization.
Comm Pure Appl Math XLIX:217--236
  
    
  
\bibitem{Bandle1980}
Bandle C (1980)
Isoperimetric inequalities and applications.
Pitman

% \bibitem{Bandle1984}
% Bandle J, Mossino J (1984)
% Rearrangements in variational inequalities.
% Ann Mat Pura e Appl (4)138:1--14

\bibitem{Barash2002}
Barash D (2002) Fundamental relationship between bilateral filtering, adaptive
  smoothing, and the nonlinear diffusion equation. IEEE T Pattern Anal
  24(6):844--847


\bibitem{Barash2004}
Barash D, Comaniciu D (2004) A common framework for nonlinear diffusion,
  adaptive smoothing, bilateral filtering and mean shift. Image Vision Comput
  22(1):73--81

  
  
\bibitem{Buades2005}
Buades A, Coll B, Morel JM (2005) A review of image denoising algorithms, with
  a new one. Multiscale Model Sim 4(2):490--530


\bibitem{Buades2006}
Buades A, Coll B, Morel JM (2006{\natexlab{a}}) Neighborhood filters and pde's.
  Numer Math 105(1):1--34


% \bibitem{Buades2006a}
% Buades A, Coll B, Morel JM (2006{\natexlab{b}}) The staircasing effect in
%   neighborhood filters and its solution. IEEE T Image Process 15(6):1499--1505


\bibitem{Buades2010}
Buades A, Coll B, Morel JM (2010) Image denoising methods. A new nonlocal
  principle. Siam Rev 52(1):113--147

%   \bibitem{Caselles2010}
%   Caselles V, Monasse P (2010) Geometric description of images as topographic maps.
%   Springer 


 
\bibitem{Diaz1992} 
D\'{\i}az JI (1992)
Symmetrization of nonlinear elliptic and parabolic problems and applications:
a particular overview.
Progress in partial differential equations elliptic and parabolic
problems, Pitman Research Notes in Mathematics, Longman, Harlow, Essex(266):
1--16

\bibitem{Diaz1995} 
D\'{\i}az JI, Nagai T (1995)
Symmetrization in a parabolic-elliptic system related to chemotaxis.
Adv Math Sci Appl 5:659--680


\bibitem{Diaz1996}
D\'{\i}az JI, Rakotoson JM (1996)
On a nonlocal stationary free boundary problem arising in the confinement of
a plasma in a Stellarator geometry.
Arch Rat Mech Anal 134(1):53--95

\bibitem{Diaz1998}
D\'{\i}az JI, Padial JF, Rakotoson JM (1998)
Mathematical treatement of the magnetic confinement in a current carrying
Stellerator.
Nonlinear Anal. TMA, 34:857--887

\bibitem{Diaz2004}
D\'{\i}az JI, Lerena MB, Padial JF, Rakotoson JM (2004)
An elliptic-parabolic equation with a nonlocal term for the transient regime of a plasma in a Stellarator.
J Differential Equations 198(2):321--355

 
% \bibitem{eisemann2004}
% Eisemann E, Durabd F (2004) Flash photography enhancement via intrinsic relighting. ACM Trans. on Graphics 23, 3. Proc. of ACM SIGGRAPH conf


\bibitem{Elad2002}
Elad M (2002) On the origin of the bilateral filter and ways to improve it.
  Ieee T Image Process 11(10):1141--1151

\bibitem{Fiorenza2007}
Fiorenza A, Rakotoson JM (2007) 
Relative rearrangement and Lebesgue spaces $L^p(\cdot)$ with variable exponent.
J Math Pures Appl 88(6):506--521

\bibitem{Fiorenza2009}
Fiorenza A, Rakotoson JM (2009) 
Relative rearrangement method for estimating dual norms.
Indiana Univ Math J  58(3):1127--1150


\bibitem{gv2013}
Galiano G, Velasco J (2013) 
On a non-local spectrogram for denoising one-dimensional signals
To appear in ACM (arXiv:1311.3269 [cs.CV])

\bibitem{gv2014}
Galiano G, Velasco J (2013) 
Neighborhood filters and the decreasing rearrangement. 
J Math Imaging Vision doi: 10.1007/s10851-014-0522-3
(arXiv:1311.2191 [cs.CV])
  
\bibitem{Gilboa2008}
Gilboa G, Osher S (2008) Nonlocal operators with applications to image
  processing. Multiscale Model Sim 7(3):1005--1028

\bibitem{Hardy1964}
Hardy GH, Littlewood JE, Polya G (1964)
Inequalities.
Cambridge University Press.
% \bibitem{Kindermann2005}
% Kindermann S, Osher S, Jones PW (2005) Deblurring and denoising of images by
%   nonlocal functionals. Multiscale Model Sim 4(4):1091--1115

% \bibitem{Kwan1999}  
%   Kwan RK-S,  Evans AC,  Pike GB (1999) MRI simulation-based evaluation of image-processing and classification methods. IEEE T Medical Imaging 18(11):1085--97

\bibitem{Lieb2001}
Lieb EH, Loss M (2001) Analysis, vol~4. American Mathematical Soc.

% \bibitem{Lions1981}
% Lions PL (1981)
% Quelques remarques sur la sym\'etrisation de Schwartz.
% Pitman Res Notes in Math 53:308--319

\bibitem{matsuo2013} 
Matsuo T, Fukushima N, Ishibashi Y (2013) Weighted Joint Bilateral Filter with Slope Depth Compensation Filter for Depth Map Refinement. 
Proc. Int Conf Computer Vision Th Appl (VISAPP 2013) 2:300--309

\bibitem{Mossino1981}
Mossino J, Temam R (1981)
Directional derivative of the increasing rearrangement mapping and application
to a queer differential equation in plasma physics.
Duke Math J 48(3):475--495


\bibitem{Mossino1984}
Mossino J (1984)
In\'egalit\'es Isop\'erm\'etriques et applications en physique.
Hermann


\bibitem{Mossino1986}
Mossino J, Rakotoson JM (1986)
Isoperimetric inequalities in parabolic equations.
Ann Sc Norm Super Pisa Sci(4) 13(1):51--73

\bibitem{Perona1990}
Perona P, Malik J (1990) Scale-space and edge detection using anisotropic
  diffusion. IEEE T Pattern Anal 12(7):629--639

%   \bibitem{petschnnig2004}
% Petschnigg G, Agrawala M, Hoppe H, Szeliski R, Cohen MF, Toyama K (2004) 
% Digital photography with flash and no-flash image pairs. 
% ACM Trans. on Graphics 23, 3. Proc. of ACM SIGGRAPH Conf.

\bibitem{Peyre2008}
Peyr{\'e} G (2008) Image processing with nonlocal spectral bases. Multiscale
  Model Sim 7(2):703--730

\bibitem{Polya1951}
P\'olya G, Szeg\"o, WN (1951)
Isoperimetric inequalities in mathematical physics.
Princenton U P

\bibitem{Rakotoson2008}
Rakotoson JM (2008) 
R{\'e}arrangement Relatif: Un instrument destimations dans 
les probl{\v{c}}mes aux limites.
Springer

\bibitem{Rakotoson2010}
Rakotoson JM (2010)
Lipschitz properties in variable exponent problems via relative rearrangement.
Chin Ann Math 31B(6):991--1006

\bibitem{Rakotoson2012}
Rakotoson JM (2012)
New Hardy inequalities and behaviour of linear elliptic equations.
J Funct Anal 263(9):2893--2920


\bibitem{Rudin1992}
Rudin LI, Osher S, Fatemi E (1992) Nonlinear total variation based noise
  removal algorithms. Physica D 60(1):259--268

\bibitem{Singer2009}
Singer A, Shkolnisky Y, Nadler B (2009) Diffusion interpretation of nonlocal
  neighborhood filters for signal denoising. SIAM J Imaging Sci 2(1):118--139.

\bibitem{Smith1997}
Smith SM, Brady JM (1997) Susan. a new approach to low level image processing.
  Int J Comput Vision 23(1):45--78

  
  
  \bibitem{Talenti1976}
Talenti G (1976)
Best constant in Sobolev inequality.
Ann Mat Pura Appli (4)110:353--372

\bibitem{Tomasi1998}
Tomasi C, Manduchi R (1998) 
Bilateral filtering for gray and color images. 
In: Sixth IEEE Int Conf Computer Vision:839--846
  
\bibitem{Vazquez1982}
  V\'azquez JL (1982)
Sym\'etrization pour $u_t=\Delta\phi(u)$ et applications, 
  C R Acad Paris 295:71--74
  
\bibitem{Yaroslavsky1985}
Yaroslavsky LP (1985) Digital picture processing. An introduction. Springer
  Verlag, Berlin

\bibitem{Yaroslavsky2003}
Yaroslavsky LP, Eden M (2003) Fundamentals of Digital Optics. Birkh\"auser,
  Boston

\end{thebibliography}
\end{document}